\documentclass[journal]{IEEEtran}
\usepackage{times}
\usepackage{graphicx}
\usepackage{bbm}
\usepackage{algorithm}
\usepackage{algorithmic}
\usepackage{multicol}
\usepackage{multirow}
\usepackage{booktabs}
\usepackage{array}
\usepackage[T1]{fontenc}
\usepackage[latin9]{inputenc}
\usepackage{cite}
\usepackage{color}
\usepackage{collcell}
\usepackage{hhline}
\usepackage{pgf}
\usepackage{pgfplots}
\usepackage{pdfpages}
\usepackage{amssymb}
\usepackage{amsmath,bm}
\usepackage{amsfonts}
\usepackage[colorlinks, linkcolor=red, anchorcolor=blue, citecolor=green]{hyperref}
\usepackage{graphicx}
\usepackage{rotating}
\usepackage{changepage}
\usepackage{makecell}
\usepackage{color, colortbl}
\usepackage{url}
\usepackage{pifont}
\usepackage{xcolor}
\usepackage{threeparttable}
\usepackage[caption=false,font=footnotesize]{subfig}

\usepackage{hanging}
\usepackage[flushmargin]{footmisc}
\newcommand{\fn}[1]{\footnote{\hangpara{1.3em}{1} #1}}

\newcommand\Tstrut{\rule{0pt}{2.6ex}}         % = `top' strut
\newcommand\Bstrut{\rule[-0.9ex]{0pt}{0pt}}   % = `bottom' strut

\ifCLASSINFOpdf
\else
\fi

\begin{document}
% \includepdf[pages=-]{revision2.pdf}

%\includepdf[pages=-]{revision.pdf}

\title{Anomaly Detection in Aerial Videos with Transformers}

\author{Pu~Jin,
    \and Lichao~Mou,
    \and Gui-Song~Xia, %~\IEEEmembership{Senior Member,~IEEE}, and
    \and  Xiao~Xiang~Zhu%,~\IEEEmembership{Fellow,~IEEE}

% \thanks{
% This work is funded by the German Federal Ministry of Education and Research (BMBF) in the framework of the international Future AI lab "AI4EO -- Artificial Intelligence for Earth Observation: Reasoning, Uncertainties, Ethics and Beyond" and jointly supported by the European Research Council (ERC) under the European Union's Horizon 2020 research and innovation programme (grant agreement No. [ERC-2016-StG-714087], Acronym: \textit{So2Sat}), by the Helmholtz Association through the Framework of Helmholtz Artificial Intelligence Cooperation Unit (HAICU) - Local Unit "Munich Unit @Aeronautics, Space and Transport (MASTr)" and Helmholtz Excellent Professorship "Data Science in Earth Observation - Big Data Fusion for Urban Research". %\textit{(Corresponding author: Xiao Xiang Zhu.)}

\thanks{This work is jointly supported by the European Research Council (ERC) under the European Union's Horizon 2020 research and innovation programme (grant agreement No. [ERC-2016-StG-714087], Acronym: \textit{So2Sat}), by the Helmholtz Association through the Framework of Helmholtz AI (grant  number:  ZT-I-PF-5-01) - Local Unit ``Munich Unit @Aeronautics, Space and Transport (MASTr)'' and Helmholtz Excellent Professorship ``Data Science in Earth Observation - Big Data Fusion for Urban Research'' (grant number: W2-W3-100), by the German Federal Ministry of Education and Research (BMBF) in the framework of the international future AI lab ``AI4EO -- Artificial Intelligence for Earth Observation: Reasoning, Uncertainties, Ethics and Beyond'' (grant number: 01DD20001) and by the German Federal Ministry for Economic Affairs and Climate Action in the framework of the ``national center of excellence ML4Earth'' (grant number: 50EE2201C).

%P. Jin is with the State Key Laboratory for Information Engineering in Surveying, Mapping and Remote Sensing (LIESMARS), Wuhan University, Wuhan 430072, China, and also with the Technical University of Munich (TUM), Germany (e-mail: pu.jin@tum.de).

P. Jin, L. Mou, and X. X. Zhu are with the Remote Sensing Technology Institute, German Aerospace Center, 82234 We{\ss}ling, Germany, and also with the Data Science in Earth Observation (former: Signal Processing in Earth Observation), Technical University of Munich, 80333 Munich, Germany. (e-mails: pu.jin@dlr.de;lichao.mou@dlr.de; xiaoxiang.zhu@dlr.de).

G.-S. Xia is with the State Key Laboratory for Information Engineering in Surveying, Mapping and Remote
Sensing (LIESMARS), and also with the School of Computer Science, Wuhan University,
Wuhan 430072, China (e-mail: guisong.xia@whu.edu.cn).}

}

\markboth{}%
{Shell \MakeLowercase{\textit{et al.}}: Bare Demo of IEEEtran.cls for Journals}

\maketitle

% \textcolor{blue}{}
\begin{abstract}
\textcolor{blue}{This work has been accepted by IEEE TGRS for publication.} Unmanned aerial vehicles (UAVs) are widely applied for purposes of inspection, search, and rescue operations by the virtue of low-cost, large-coverage, real-time, and high-resolution data acquisition capacities. Massive volumes of aerial videos are produced in these processes, in which normal events often account for an overwhelming proportion. It is extremely difficult to localize and extract abnormal events containing potentially valuable information from long video streams manually. Therefore, we are dedicated to developing anomaly detection methods to solve this issue. In this paper, we create a new dataset, named Drone-Anomaly, for anomaly detection in aerial videos. This dataset provides 37 training video sequences and 22 testing video sequences from 7 different realistic scenes with various anomalous events. There are 87,488 color video frames (51,635 for training and 35,853 for testing) with the size of 640 $\times$ 640 at 30 frames per second. Based on this dataset, we evaluate existing methods and offer a benchmark for this task. Furthermore, we present a new baseline model, ANomaly Detection with Transformers (ANDT), which treats consecutive video frames as a sequence of tubelets, utilizes a Transformer encoder to learn feature representations from the sequence, and leverages a decoder to predict the next frame. Our network models normality in the training phase and identifies an event with unpredictable temporal dynamics as an anomaly in the test phase. Moreover, To comprehensively evaluate the performance of our proposed method, we use not only our Drone-Anomaly dataset but also another dataset. We will make our dataset and code publicly available. A demo video is available at \url{https://youtu.be/ancczYryOBY}. We make our dataset and code publicly available\fn{\url{https://gitlab.lrz.de/ai4eo/reasoning/drone-anomaly} \\ \url{https://github.com/Jin-Pu/Drone-Anomaly}}. 

%release the code and dataset at \href{https://gitlab.lrz.de/ai4eo/reasoning/drone-anomaly}{GitLab} and \href{https://github.com/Jin-Pu/Drone-Anomaly}{GitHub}. 

% most anomaly detection methods adopt an autoencoder architecture and utilize a convolution-based encoder. We propose a new network termed ANDT, which utilizes an attention-based Transformer encoder. Specifically, we interpret a video as a sequence of patches and feed them into a Transformer encoder that integrates information among all patches to learn the global context. A simple decoder is then combined with the encoder for predicting the next frame based on the learned spatiotemporal dependencies. Moreover, we conduct massive ablation studies and experiments for validating the effectiveness of our approach. 

% However, this encoder is unable to effectively capture long-term temporal dependencies among input frames due to limited temporal receptive fields. We observe that a video can be naturally regarded as a temporal sequence. Hence, for learning long-term temporal relations, we treat a video into a sequence of patches and feed them into an attention-based Transformer encoder that integrates information among all patches to learn the global context. A simple decoder is then combined with the encoder for predicting the next frame based on the learned spatiotemporal dependencies. In addition, our network learns the normality in the training phase and identifies an event with unpredictable temporal dynamics as an anomaly in the test phase. Moreover, we conduct massive ablation studies and experiments for validating the effectiveness of our approach.

\end{abstract}

\begin{IEEEkeywords}
Anomaly detection, aerial videos, convolutional neural networks (CNNs), Transformers, temporal reasoning, unmanned aerial vehicle (UAV).
\end{IEEEkeywords}

\IEEEpeerreviewmaketitle

\section{Introduction}

\begin{figure}[]
	\centering
 	\includegraphics[width = 1\linewidth]{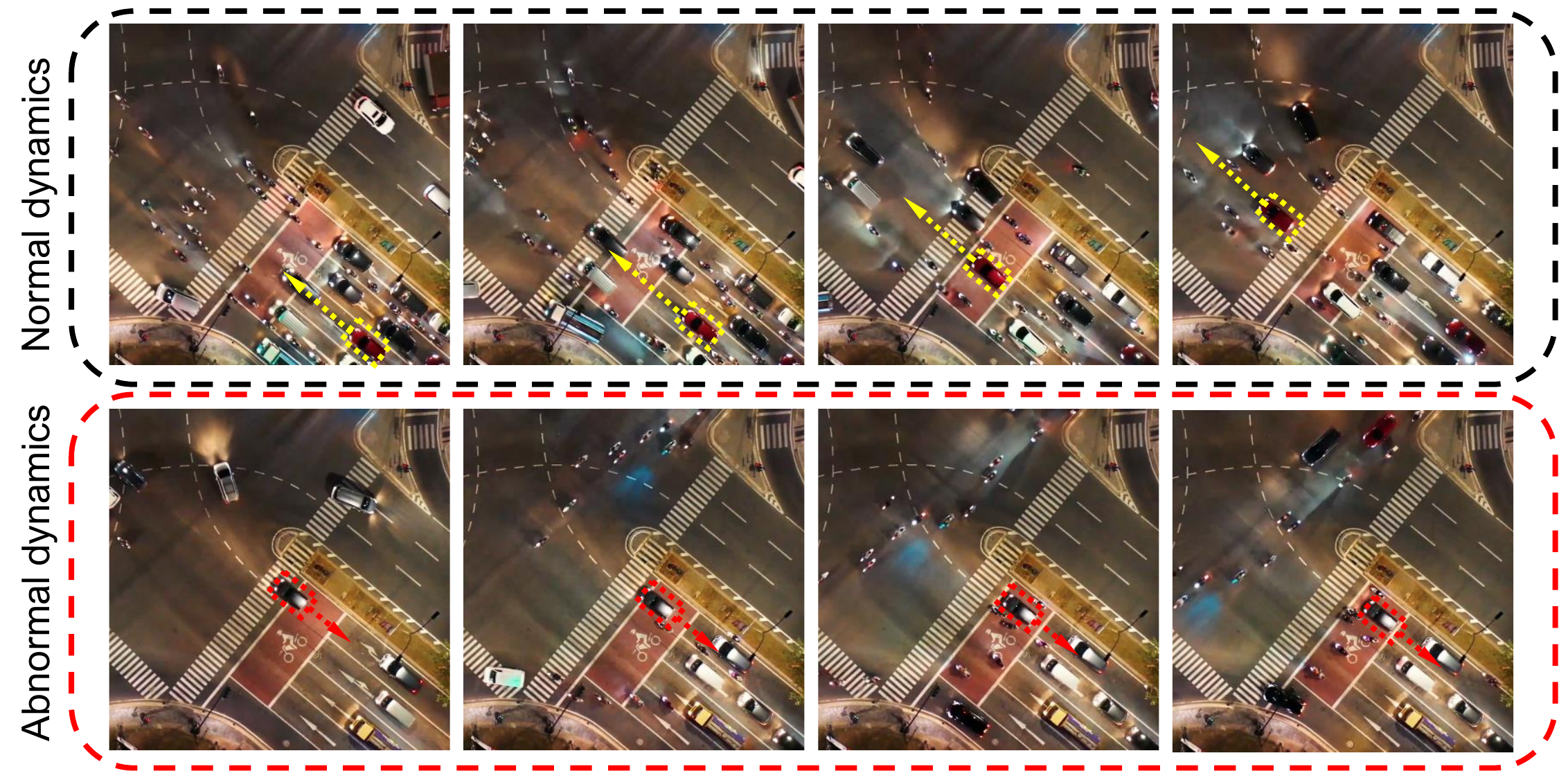}
	\caption{\textbf{Abnormal and normal dynamics.} We display some frames from the \emph{crossroads} scene for demonstrating the importance of temporal information in detecting anomalous events in aerial videos. In the normal video clip (top), all vehicles have a consistent moving direction. We use a \textcolor{yellow}{yellow} box with an arrow to represent an example vehicle and its moving direction. In the abnormal video snippet (bottom), a vehicle (in the \textcolor{red}{red} box) moves backward on the road. We can see the importance of temporal context in this task.}
	\label{cover}
\end{figure}

\IEEEPARstart{A}{nomaly} detection refers to the detection of visual instances that significantly deviate from the majority~\cite{pang2021deep}. Due to the expanding demand in broad domains, such as inspection~\cite{xie2020visualised,neto2012validation,zhou2019anomalynet,au2006anomaly,tariq2021anomaly}, search operations~\cite{morse2012color,eismann2009automated}, and security~\cite{garg2020multi,yang2019nlsalog,yang2017anomaly,desnitsky2015detection}, anomaly detection plays increasingly important roles in various communities including computer vision, data mining, machine learning, and remote sensing. With the proliferation of UAVs worldwide, massive produced aerial videos spur the demand for detecting abnormal events in aerial video sequences in a wide range of applications~\cite{mou2020era}. For example, many long-endurance UAVs\footnote{\url{https://www.airforce-technology.com/features/featurethe-top-10-longest-range-unmanned-aerial-vehicles-uavs/}} are developed and utilized in inspection operations\cite{xie2020visualised,neto2012validation,zhou2019anomalynet,au2006anomaly,tariq2021anomaly}. Large amounts of aerial videos are created by these UAVs, in which normal video segments often account for an overwhelming proportion of the whole video. It is time-consuming and costly to find potentially valuable information from long and untrimmed videos manually. Therefore, we are intended to adopt anomaly detection methods to temporally localize anomalous events in aerial videos automatically.

\begin{figure*}[h]
	\centering
 	\includegraphics[width = 1\linewidth]{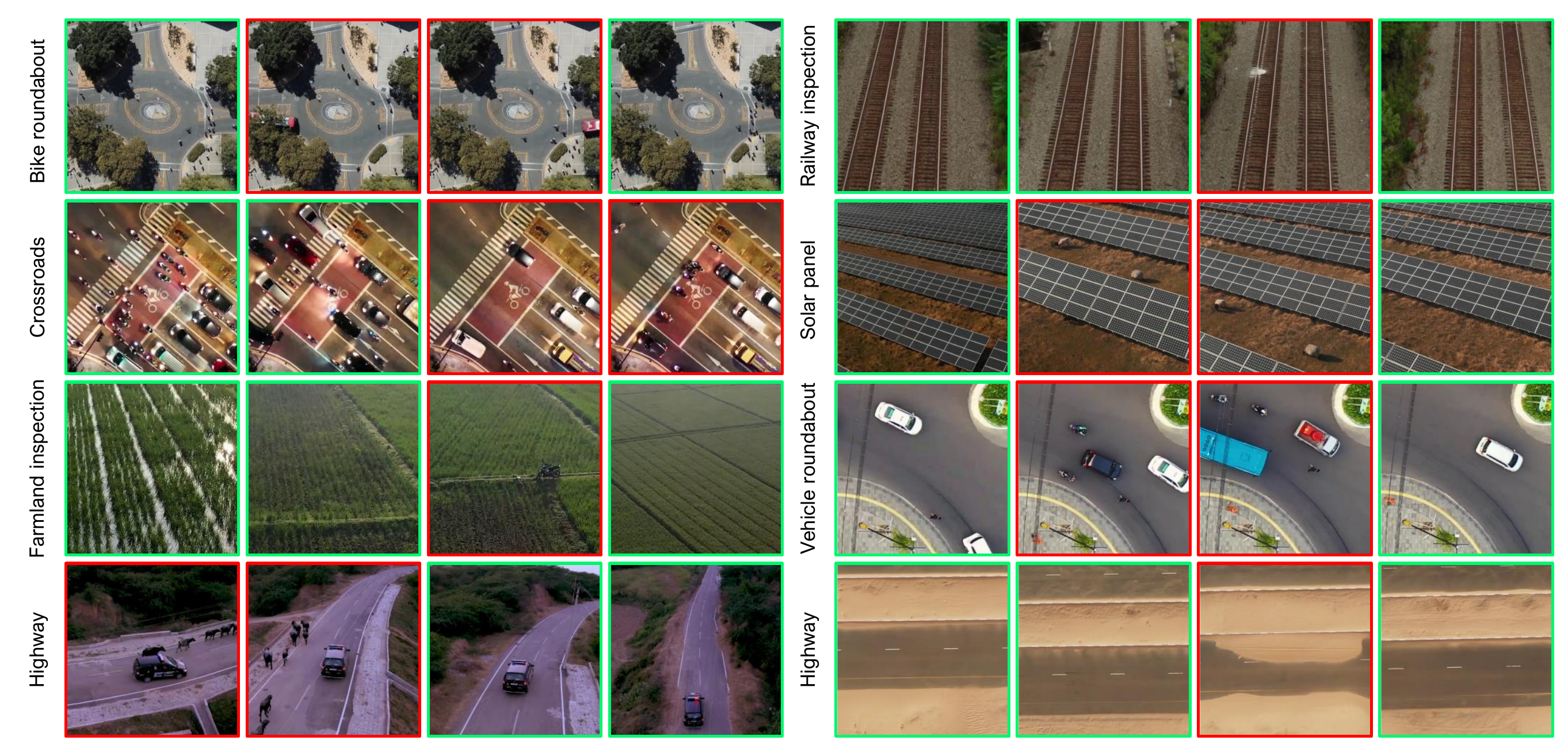}
	\caption{\textbf{Overview of the Drone-Anomaly dataset.} We show four frames of each video. The anomalous frames are marked with \textcolor{red}{red} borders, and frames with \textcolor{green}{green} borders are normal ones.}
	\label{dataset_samples}
\end{figure*}

% We cannot know beforehand what anomalous events are in a scene, because there are too many possibilities that are impossible to be exhaustively listed. On the other hand, it is easy to figure out what is normal in the scene.

% Anomalies cannot be known beforehand, because there are too many possibilities that are impossible to be exhaustively listed. In contrast, it is easy to figure out the normality.

Usually we cannot know beforehand what anomalies are in a scene, because there are too many possibilities that are impossible to be exhaustively listed. By contrast, it is easy to have information on the nature of normality in advance. Hence, most existing methods for anomaly detection only use normal data to learn feature representations of normality and consider test instances that cannot be well described as anomalies. Massive studies~\cite{kiani2020optimal,dias2020incongruence,liu2017unsupervised,wang2019automatic, yang2018ship, stein2002anomaly, chang2002anomaly, kang2017hyperspectral, yuan2015hyperspectral, liu2021abnet, wang2021auto, zhang2021spectral, xie2021weakly} are dedicated to detecting and categorizing non-conforming patterns present in images. These studies mainly focus on spatial occurrences of anomalous patterns. In contrast, anomaly detection in videos aims at identifying temporal occurrences (i.e., start and end times) of abnormal events. In computer vision, many methods~\cite{golan2018deep,sabokrou2018adversarially,zhai2016deep,zimek2012survey,zong2018deep, lin2021learning, salehi2021multiresolution, ristea2022self} have been proposed for this task in surveillance videos. In comparison with surveillance videos, UAV videos bring the following challenges: (1) moving-camera instead of static-camera; (2) variable spatial resolution due to changes in flight altitude. Existing works~\cite{cavaliere2018empowering, yang2021proactive, cavaliere2019human} predefine several categories of anomalous events, convert aerial video anomaly detection into an event recognition task, and utilize supervised methods to address this problem. By contrast, in this work, we are interested in unsupervised methodologies for this task. Because in many real-world applications, it is not possible to exhaustively list all anomalous events beforehand. More specifically, we train a model for anomaly detection in aerial videos using only normal data that can be collected easily in advance.

% due to several constraints related to the acquisition platform and aerial video data such as (1) variation in the camera altitude, in point of view, (2) various video scenes with the dynamic and complicated backgrounds, (3) lack of public datasets for anomaly detection in aerial videos.

% And video sequence involves complicated high-dimensional spatiotemporal representations~\cite{pang2021deep}. Therefore, reasoning different dependencies over appearance and motion features are needed to ensure a reliable prediction of an anomaly in a video sequence. 

% which expands the scope of video anomaly detection research. 

In this paper, we focus on detecting anomalous events in aerial videos. To this end, we create a new dataset, named Drone-Anomaly, providing $37$ training video sequences and $22$ testing video sequences from $7$ different realistic scenes. The dataset contains real-world anomalous events that are not staged by actors. Based on this dataset, we evaluate existing methods and offer a benchmark. In addition, we note that modeling temporal context is critical (see Fig.~\ref{cover}). Most existing anomaly detection methods utilize convolution-based encoders for capturing spatiotemporal dependencies among the input video frames. However, this is limited in learning long-term relations due to limited temporal receptive fields of these models. In this paper, we present a new baseline model, ANomaly Detection with Transformers (ANDT), which takes as input several consecutive video frames, leverages a Transformer encoder to model global context, and utilizes a decoder to predict the next frame. More specifically, ANDT treats a video as a sequence of tubelets and maps them into tubelet embeddings by linear projection. For preserving spatiotemporal information, the tubelet embeddings are added with learnable spatiotemporal position embeddings and then fed into a Transformer encoder to learn a spatiotemporal feature. The decoder is subsequently combined with the encoder for predicting the next frame based on the learned spatiotemporal representation. Our network is able to well predict an event with normal temporal dynamics and identifies an event with unpredictable temporal dynamics as an anomaly in the test phase.

% Moreover, we observe that a video can be naturally considered as a temporal sequence.

The main contributions of this paper can be summarized as follows:

\begin{itemize}
	
    \item We create an annotated dataset consisting of $37$ training videos and $22$ testing videos involving $7$ realistic scenes, covering a large variety of anomalous events. This dataset expands the scope of anomaly detection research. In addition, we extensively validate existing methods in order to provide a benchmark for this task.
%. Specifically, it treats a video as a sequence of patches and leverages a Transformer encoder to learn a spatiotemporal feature among the patch embeddings. And a decoder is then combined with the encoder for predicting the next frame based on the learned spatiotemporal representation. And we

    \item We extensively validate existing methods in order to provide a challenging benchmark for anomaly detection in aerial videos.
    
    \item We present a new baseline model ANDT and conduct extensive ablation studies and experiments for validating the effectiveness of our approach. To the best of our knowledge, this is the first time that a Transformer-based network is proposed for video anomaly detection.

\end{itemize}

The remaining sections of this paper are organized as follows. The related works are introduced in Section~\ref{Related Works}. Then, we detail our new dataset in Section~\ref{sec:Dataset}. And our network is described in Section~\ref{sec:Methodology}. Section~\ref{sec:Experiments} shows and discusses experimental results. Finally, the paper is concluded in Section~\ref{Conclusion}. 
\begin{table*}[]
\caption{\textbf{Dataset details.} We provide variable details of the Drone-Anomaly dataset.}
\label{dataset_details}
\centering
\begin{tabular}{c|ccc}
\toprule
\textbf{}      Scene        & \# Video snippets (Train / Test)  & \# Frames (Train / Test) & Example anomalies \\
\midrule
\midrule
Highway   & 6 / 3  & 9045 / 2820 & Animals walking on the street; Car collision \\ [4pt]
Crossroads   & 10 / 5  & 15772 / 6244  & Retrograde vehicles; Traffic congestion  \\ [4pt]
Bike roundabout   & 6 / 7  & 7950 / 18427 & Moving vehicles  \\ [4pt]
Vehicle roundabout   & 4 / 2  & 5266 / 2643 & People crossing the road  \\ [4pt]
Railway inspection   & 3 / 1  & 1206  /  882 & Obstacles on the railway  \\ [4pt]
Solar panel inspection   & 4 / 3  & 2848 / 2450 & Unknown objects; Defects of panel
  \\ [4pt]
Farmland inspection   & 4 / 1  & 9548 / 2387 & Unidentified vehicles  \\

\bottomrule
\end{tabular}
\end{table*}

\begin{table*}[]
\caption{\textbf{Comparison with related datasets.} We offer various comparisions for each datasets.}
\label{dataset_details}
\centering
\begin{threeparttable}
\begin{tabular}{c|cccccc}
\toprule
\textbf{}      Dataset   & \# Videos  & \# Frames & \# Scenes  & Type of task & Type of anomalies & Year\\
\midrule
\midrule
Mini-drone~\cite{bonetto2015privacy}   & 38 & 22,860  & 1 & Event recognition and detection & Actor-staged anomalies & 2015\\ [4pt]
AU-AIR-Anomaly\tnote{*}~\cite{bozcan2021context}  & 1 & 32,823  & 1 & Anomaly detection & Realistic anomalies & 2021\\ [4pt]
Drone-Anomaly   & 59 & 87,488  & 7 & Anomaly detection & Realistic anomalies & 2022\\ 
\bottomrule
\end{tabular}
\begin{tablenotes}
    \item[*] The AU-AIR dataset is originally created for object detection tasks. 
\end{tablenotes}
\end{threeparttable}
\end{table*}

\section{Related Work} \label{Related Works}

% \begin{table*}[]
% \caption{\textbf{Dataset details.} We provide variable details of the Drone-Anomaly dataset.}
% \label{dataset_details}
% \centering
% \begin{tabular}{@{}c|cccc@{}}
% \toprule
% \textbf{}      Scene        & \# Video snippets (Train / Test)  & \# Frames (Train / Test) & Total duration & Example anomalies \\
% \midrule
% \midrule
% Highway   & 6 / 3  & 9045 / 2820  & 395.50s & Animals walking on the street; Car collision \\ [4pt]
% Crossroads   & 10 / 5  & 15772 / 6244  & 733.87s & Retrograde vehicles; Traffic congestion  \\ [4pt]
% Bike roundabout   & 6 / 7  & 7950 / 18427  & 879.23s & Moving vehicles  \\ [4pt]
% Vehicle roundabout   & 4 / 2  & 5266 / 2643  & 263.63s & People crossing the road  \\ [4pt]
% Railway inspection   & 3 / 1  & 1206  /  882  & 69.60s & Obstacles on the railway  \\ [4pt]
% Solar panel inspection   & 5 / 4  & 3016 / 2629  & 188.17s & Unknown objects; Defects of panel
%   \\ [4pt]
% Farmland inspection   & 4 / 1  & 9548 / 2387  & 397.83s & Unidentified vehicles  \\

% \bottomrule
% \end{tabular}
% \end{table*}

In remote sensing, there have been a number of works for anomaly detection in hyperspectral imagery~\cite{reed1990adaptive,li2014collaborative,chen2014deep,hu2015deep,matteoli2010tutorial,banerjee2006support,du2010random,kwon2005kernel,schaum2007hyperspectral}. These studies mainly focus on locating pixels with significantly different spectral signatures from their neighboring background pixels in the spatial domain. For example, the Reed-Xiaoli (RX) algorithm~\cite{reed1990adaptive} uses local Gaussian model to detect anomalies in hyperspectral images and has become a baseline model. In~\cite{li2014collaborative}, a collaborative representation detector (CRD) is proposed to detect pixels with unknown spectral signatures. Recently, deep learning-based methods have drawn significant attention.~\cite{chen2014deep} proposes to use an autoencoder to learn representative features in order to detect anomalies in an unsupervised manner. In~\cite{hu2015deep}, the authors employ convolutional neural networks (CNNs) to learn spectral-spatial features in this task and achieve outstanding performance.

From static imagery to multi-temporal images, much effort~\cite{shaw2005evaluating,liu2021detection,chemura2017integrating,lasaponara2006use,lanorte2015use,malhotra2015long,he2019temporal,zhang2021unsupervised} has been made to detect anomalies in the temporal domain. For instance,~\cite{shaw2005evaluating} uses multispectral images over two years for locating and identifying crop anomalies in two soybean fields. In~\cite{liu2021detection}, the authors leverage multi-temporal thermal infrared (TIR) images for detecting geothermal anomaly areas by spatiotemporal analysis. In~\cite{chemura2017integrating}, multi-temporal Landsat images are utilized to detect NDVI anomalies for mapping incongruous patches in coffee plantations.

Moreover, we notice that in computer vision, many anomaly detection approaches~\cite{chong2017abnormal,hasan2016learning,luo2017revisit,ravanbakhsh2017abnormal, zhao2017spatio,luo2017remembering, medel2016anomaly} have been developed for fixed camera surveillance videos. By contrast, we think that anomaly detection in aerial videos is more challenging because the videos are usually acquired by moving cameras. There have been a few works for investigating anomaly detection in aerial videos. These works~\cite{cavaliere2018empowering, yang2021proactive, cavaliere2019human} regard this problem as an event recognition task. Specifically, they first predefine several anomalous activities and then leverage supervised methods to recognize the defined events from aerial videos. For example, ~\cite{cavaliere2018empowering} leverages object tracking and classification methods to obtain trajectories and semantic information, then utilizes an ontology-based reasoning model to learn spatiotemporal relations among them for detecting video events. In~\cite{yang2021proactive}, the authors define three different safety-related anomalies and propose a functional approach that models temporal relations of time-to-collision safety indicators to detect these anomalies from UAV-based traffic videos. Furthermore,~\cite{cavaliere2019human} proposes a hybrid approach that integrates trajectories and semantic information of objects to build high-level knowledge for extracting complicated critical activities and events from UAV videos. Most recently, based on AU-AIR dataset~\cite{bozcan2020air} that is proposed for object detection in UAV videos,~\cite{bozcan2021context} builds a dataset including several anomalous objects (hereafter we call it AU-AIR-Anomaly dataset) and proposes a supervised method CADNet to detect instances and contextual anomalies in aerial videos. Compared to our dataset, the AU-AIR-Anomaly dataset only contains a single scene, i.e., traffic, and its aerial video has a relatively stable perspective.

In real-world applications, there are many possible anomalies existing in a scenario, which cannot be exhaustively listed and defined in advance. Instead, the nature of normality is relatively stable and easy to know beforehand. Therefore, we propose an unsupervised method ANDT that learns feature representations of genetic normality from merely normal data and determines test data with large reconstruction errors as anomalies. Moreover, methods~\cite{cavaliere2018empowering, yang2021proactive, cavaliere2019human, jin2021anomaly, bozcan2021context} all leverage convolution-based encoders for learning spatiotemporal dependencies among input video frames. Due to the limited temporal receptive fields, these models are unable to effectively capture long-term temporal relations. By contrast, our method ANDT adopts a Transformer-based encoder that confers our model with a global temporal receptive field and enables it to capture temporal dependencies among all input frames. With a global perspective, our model is adept at distinguishing the movement of instances from the dynamic background and provides rich contextual information for detecting anomalies.

\section{Dataset}
\label{sec:Dataset}

To address the lack of available datasets for anomaly detection in aerial videos, we present the Drone-Anomaly. This section introduces the construction of our dataset, including video collection and annotation. Finally, we present the overall statistics of the dataset.

\subsection{Video Collection}

We collect aerial videos on YouTube\footnote{\url{https://www.youtube.com/}} and Pexels\footnote{\url{https://www.pexels.com/}} using search queries (e.g., \emph{drone highway}, \emph{UAV roundabout}) for each scene. In order to increase the diversity of anomalous events, we retrieve aerial videos using different languages (e.g., English, Chinese, German, and French). Moreover, to ensure the quality of aerial videos, we remove videos with any of the following situations: too short duration, manually edited, not captured by UAV cameras, without clear anomalous events. We show four frames of an example video from each scene in Fig.~\ref{dataset_samples}.

\subsection{Annotation}

We assign video-level labels for training data. In the test phase, frame-level annotations are needed to evaluate the performance. Thus, we provide frame-level labels with binary values, where anomalous frames are labeled as 1, and 0 indicates normal frames. For each scene, training videos and testing videos with anomalies are provided. The details are shown in Table~\ref{dataset_details}.

\subsection{Statistics}

Our Drone-Anomaly dataset consists of long, untrimmed aerial videos that cover $7$ real-world scenes, including \emph{highway}, \emph{crossroads}, \emph{bike roundabout}, \emph{vehicle roundabout}, \emph{railway inspection}, \emph{solar panel inspection}, and \emph{farmland inspection}. Various anomalies in these scenes have important practical significance and applications. We provide the overview of our dataset in Table~\ref{dataset_details}. Basically, the dataset consists of $37$ training video sequences and $22$ testing sequences. Each of them is at $30$ frames per second and with a spatial size of $640 \times 640$ pixels. There are a total of $87,488$ color video frames ($51,635$ for training and $35,853$ for testing). 

\subsection{Comparison with Related Datasets}

We compare our dataset with related datasets in Table~\ref{dataset_details}. Mini-drone dataset~\cite{bonetto2015privacy} consisting of 38 videos is proposed to parse video contents for privacy protection. The dataset contains three categories: normal, suspicious, and illicit behaviors. All events are staged by actors. This dataset can be used for different tasks, e.g., action recognition, video classification, event recognition, and event detection. In addition, based on the AU-AIR dataset~\cite{bozcan2020air}, ~\cite{bozcan2021context} annotates different anomalous events for detecting anomalies in aerial videos. The AU-AIR-Anomaly dataset contains four realistic anomalies, i.e., \emph{a car on a bike road}, \emph{a person on a road}, \emph{a parked van in front of a building}, and \emph{a bicycle on a road}.

\section{Methodology}
\label{sec:Methodology}

In this section, we detail our model. First, we introduce future frame prediction---the framework we use for anomaly detection, in Section~\ref{subsec:Methodology1}. Next, we give the detailed description of ANDT in Section~\ref{subsec:Methodology2}.

\subsection{Future Frame Prediction for Anomaly Detection}\label{subsec:Methodology1}
For anomaly detection in aerial videos, comparing to the commonly used reconstruction-based framework ~\cite{an2015variational,masci2011stacked,fowler2011anomaly,zong2018deep,nguyen2019anomaly,wang2020advae,akcay2018ganomaly,akccay2019skip,gong2019memorizing,park2020learning} where target values are equal to the inputs, it is more natural to predict the next video frame conditioned on several consecutive frames and compare the predicted one with its ground truth. In this way, temporal context can be modeled. The assumption of the future frame prediction framework is that temporal consistency in normal events is maintained stably, thus normal events are temporally more predictable than anomalies. In the training stage, a network is trained with only normal videos to learn normal temporal patterns. In the test phase, events and activities not perfectly predicted by the network are then deemed as anomalies.
% $\mathcal{V}=\left \{ \mathbf{I}_1, \mathbf{I}_2,...,\mathbf{I}_T \right \}$
Formally, given a video $\mathcal{V}$ composed of consecutive $T$ frames, $\mathcal{V}=\left \{ \bm {I}_1, \bm {I}_2,...,\bm {I}_T \right \}$. All frames are stacked temporally and then utilized to predict the next frame $\bm{I}_{T+1}$. The predicted frame is denoted as $\bm{\hat{I}}_{T+1}$. We aim to learn a mapping $\mathcal{P}$ as follows:

\begin{equation}
    {\mathcal{P}(\mathcal{V})\to \bm {\hat{I}}_{T+1}\,.}
\end{equation}

To make $\bm{\hat{I}}_{T+1}$ closer to $\bm{I}_{T+1}$, we minimize their $\ell_2$ distance in intensity space as follows:

\begin{equation}
    {L(\bm{\hat{I}}, \bm{I}) = \left \| \bm{\hat{I}} - \bm{I} \right \|^2_2 \,.}
\end{equation}
In the test phase, the $\ell_2$ distance between the predicted next frame $\bm{\hat{I}}_{T+1}$ and the true next frame $\bm{I}_{T+1}$ is calculated for identifying anomaly. The frames with relatively large $\ell_2$ distances are deemed as anomalies.

\begin{figure*}[]
	\centering
 	\includegraphics[width = 1\linewidth]{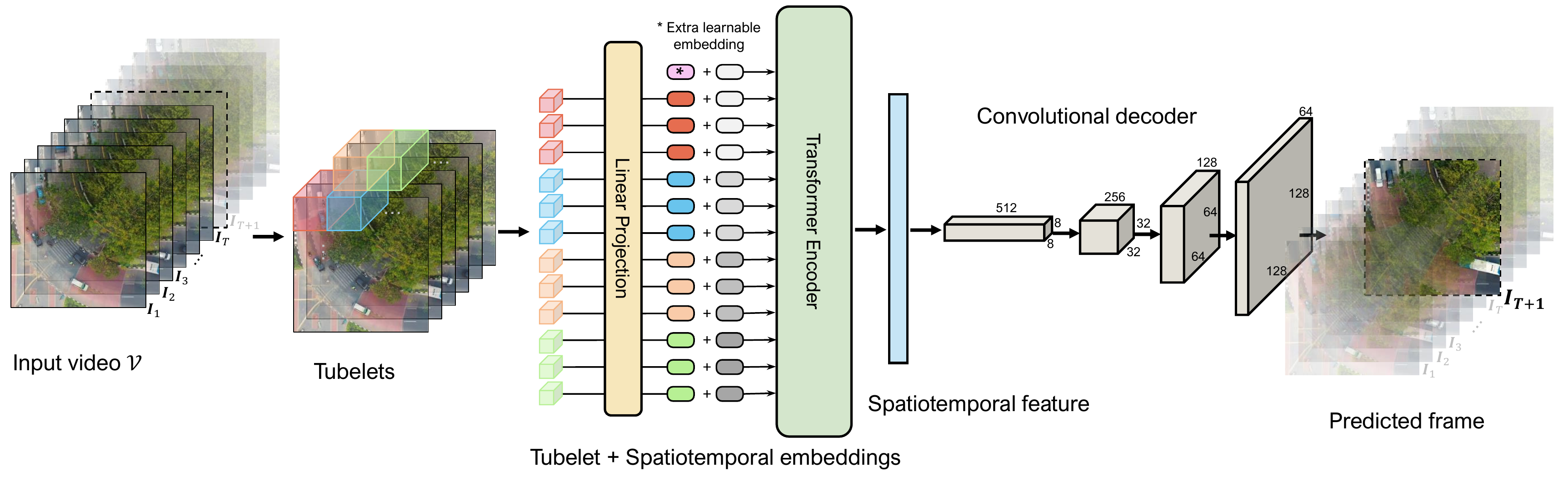}
	\caption{\textbf{The overview of ANDT.} Our method treats a video as a sequence of tubelets and maps them into tubelet embeddings by linear projection. For preserving spatiotemporal information, the tubelet embeddings are added with learnable spatiotemporal position embeddings and then fed into a Transformer encoder to learn a spatiotemporal feature. The decoder is subsequently combined with the encoder for predicting the next frame based on the learned spatiotemporal representation.}
	\label{overview}
\end{figure*}

\subsection{Anomaly Detection with Transformers (ANDT)}\label{subsec:Methodology2}

We propose a method ANDT as the mapping $\mathcal{P}$. The Transformer~\cite{vaswani2017attention} was originally proposed for sequence-to-sequence tasks in natural language processing (NLP), such as language translation. Its main idea is to use self-attention that enables the model to capture long-range dependencies in a whole sequence. We observe that a video is naturally a temporal sequence, but with spatial content. Therefore, we interpret a video as a sequence of tubelets and process them by a Transformer encoder in order to capture long-term spatiotemporal dependencies. Furthermore, a 3D convolutional decoder is further attached for predicting the next frame based on the learned spatiotemporal relations. An overview of the model is depicted in Fig.~\ref{overview}.

%The standard Transformer receives a 1D sequence as the input. Following the operations in Vision Transformer that processes a 2D images into 1D sequence, we handle the frames in $3D$ videos by the same ways.We first extract $i^{\text{th}}$ frame $\mathbf{I}_i\in \mathbb{R}^{H\times W\times C}$ from the video $\mathbf{V}\in \mathbb{R}^{T\times H\times W\times C}$ and reshape the frame $\mathbf{I}_i$ into a sequence of flattened 2D patches as follows:

% \begin{equation}
%     {\mathbf{x}_{k}\in \mathbb{R}^{(n_t \cdot n_h \cdot n_w) \times(t \cdot h \cdot w \cdot C)} \,,}
% \end{equation}

Vision Transformer~\cite{dosovitskiy2020image} performs tokenization by splitting an image into a sequence of small patches. In this work, since we deal with videos, we tokenize a video by extracting non-overlapping, spatiotemporal tubes. Specifically, the input video $\mathcal{V}\in \mathbb{R}^{T\times H\times W\times C}$ is split into a sequence of flattened 3D tubelets $\bm{x}_{k}\in \mathbb{R}^{(n_t \cdot n_h \cdot n_w) \times(t \cdot h \cdot w \cdot C)}$, where $(H,W)$ is the spatial size of video frames, $C$ represents the number of channels, $T$ denotes the number of frames, $(t, h, w)$ is the dimension of each tubelet, $n_t = \left [ \frac{T}{t} \right ]$, $n_h = \left [ \frac{H}{h} \right ]$, and $n_w = \left [ \frac{W}{w} \right ]$. $N = n_t \cdot n_h \cdot n_w$ is the number of tokens. Then, we map the tubelets into a $K$-dimensional latent space by a trainable linear projection with weights $\bm{E} \in \mathbb{R}^{(t \cdot h \cdot w \cdot C)\times K}$. By doing so, the spatiotemporal information can be preserved during the tokenization.

We also prepend a learnable embedding $\bm{x}_{cls}$ to the sequence of tubelet embeddings. It also serves as the output feature $\bm{p}$ of the Transformer encoder. Furthermore, to inject original spatiotemporal position information into our model, we add learnable spatiotemporal position embeddings $\bm{E}_{pos} \in \mathbb{R}^{(N + 1)\times K}$ to the tubelet embeddings. The equations are shown as follows: 

% We also investigate different position embeddings in ablation studies in Section~\ref{sec:Experiments}.
% \begin{equation}
% \begin{split}
%     % \mathbf{z}_0  = & [\mathbf{I}_{class};\mathbf{I}_{1,1}\mathbf{E};\mathbf{I}_{2,1}\mathbf{E};...;\mathbf{I}_{N,1}\mathbf{E}; \\ & ...; \mathbf{I}_{p,i}\mathbf{E};...;\mathbf{I}_{N,T}\mathbf{E}]+\mathbf{E}_{pos}\,,
%     & \bm{z}_0 = [\bm{x}_{cls};\bm{x}^{1}_{k}\bm{E};\bm{x}^{2}_{k}\bm{E};...; \bm{x}^{N}_{k}\bm{E}] + \bm{E}_{pos}\,, \\ &
%     \bm{E} \in \mathbb{R}^{(t \cdot h \cdot w \cdot C)\times K}\,, \bm{E}_{pos} \in \mathbb{R}^{(N + 1)\times K}\,.
% \end{split}
% \end{equation}
\begin{equation}
    \bm{z}_0 = [\bm{x}_{cls};\bm{x}^{1}_{k}\bm{E};\bm{x}^{2}_{k}\bm{E};...; \bm{x}^{N}_{k}\bm{E}] + \bm{E}_{pos}\,.
\end{equation}

$\bm{z}_0$ is subsequently fed into Transformer encoder layers, each consisting of two sublayers. The first is a multi-head self-attention (MSA) mechanism, and the second is a simple multilayer perceptron (MLP). Layer normalization (LN) is applied before every sublayer, and residual connections are used in every sublayer. The Transformer encoder takes as input these embeddings and learns a spatiotemporal feature $\bm{p}$ via:

% \begin{equation}
% \bm{z}_0 = [\bm{x}_{cls};\bm{x}^{1}_{k}\bm{E};\bm{x}^{2}_{k}\bm{E}; ..., \bm{x}^{N}_{k}\bm{E}] + \bm{E}_{pos}
% \end{equation}

% \begin{equation}
%     {\bm{E} \in \mathbb{R}^{(t \cdot h \cdot w \cdot C)\times K}\,, \bm{E}_{pos} \in \mathbb{R}^{(N + 1)\times K}\,,}
% \end{equation}

\begin{equation}
    {\bm{z}^{\prime}_{l}= {\rm MSA} ({\rm LN}(\bm{z}_{l-1})) + \bm{z}_{l-1}\,,}
\end{equation}
%  \quad (l = 1,\cdots,L)
\begin{equation}
    {\bm{z}_{l}= {\rm MLP}({\rm LN}({\bm z}^{\prime}_{l})) + {\bm z}^{\prime}_{l}\,,}
\end{equation}
\begin{equation}
    {\bm{p}=  {\rm LN}(\bm{z}^{0}_{L})\,,}
\end{equation}
where $l = 1,\cdots,L$.
%, and $\bm{p}$ is the spatiotemporal feature learned by the Transformer encoder. 

We leverage a convolutional decoder to predict the next frame $\bm{\hat{I}}_{i+1}$ based on the learned spatiotemporal feature $\bm{p}$. First, we leverage two fully-connected layers to increase the dimension of $\bm{p}$ and then reshape it into a 3D tensor of $8\times8\times512$. This size is associated with the number of convolutional layers in the decoder. Considering both computational complexity and reconstruction accuracy, we use an decoder with $5$ convolutional layers and upsampling layers. It progressively reconstructs the next frame with the size of $256\times256\times3$ from the encoded feature tensor of $8\times8\times512$. In particular, we leverage a progressive upsampling strategy that utilizes upsampling layers and convolution layers alternately. The upsampling rate is restricted to $2\times$. The batch normalization and ReLU are applied after each convolution layer. This strategy enables our decoder to learn spatial dependencies and upsample the learned features in a progressive manner, which leads to a better reconstruction of details and boundaries.
% \begin{equation}
%     {\mathbf{\hat{I}}_{i+1}=ReLU(Convolution(MLP(\mathbf{y}))).}
% \end{equation}

% \newcommand{\maytab}{% Just for this example
% 		\begin{tabular}{p{0.9cm}<{\centering}|p{0.6cm}<{\centering}p{0.6cm}<{\centering}p{0.6cm}<{\centering}}

%           & AUC & F1 & OA\\
%         \midrule
%         \midrule
%         $8 \times 8$ & 60.54 & 57.13 & 53.93\\
%         $16 \times 16$ & \textbf{64.32} & 63.51 & \textbf{61.56}\Bstrut\\ 
%         $32 \times 32$ & 62.78 & 59.46 & 57.20\Bstrut\\ 
%         $64 \times 64$ & 64.07 & \textbf{65.78} & 60.14\\ 
%         \end{tabular}
% 		\caption{\textbf{Patch size.} In general, the model with $16 \times 16$ exhibits superior performance and effectively preserves the spatiotemporal information of the input video.}
% 		\label{tab:label subtable A}
% }

% \newcommand{\mytable2}{% Just for this example
%         \centering
% 		\begin{tabular}{p{0.9cm}<{\centering}|p{0.6cm}<{\centering}p{0.6cm}<{\centering}p{0.6cm}<{\centering}}

%           & AUC & F1 & OA\\
%         \midrule
%         \midrule
%         1 & 62.41 & 58.32 & 56.06\Bstrut\\ 
%         2 & \textbf{67.48} & \textbf{68.53} & \textbf{63.29}\Bstrut\\ 
%         4 & 64.32 & 63.51 & 61.56\\ 
%         6 & 63.71 & 60.46 & 59.53\\
%         \end{tabular}
% 		\caption{\textbf{Number of Transformer layers.} The network with 2 layers achieves a better performance and also has relatively small computational complexity.}
% 		\label{tab:label subtable B}
% }

\newcommand{\aaa}{% Just for this example
  \begin{tabular}{p{0.9cm}<{\centering}|p{0.6cm}<{\centering}p{0.6cm}<{\centering}p{0.6cm}<{\centering}}

          & AUC & F1 & OA\\
        \midrule
        \midrule
        $8 \times 8$ & 60.54 & 57.13 & 53.93\\
        $16 \times 16$ & \textbf{64.32} & 63.51 & \textbf{61.56}\Bstrut\\ 
        $32 \times 32$ & 62.78 & 59.46 & 57.20\Bstrut\\ 
        $64 \times 64$ & 64.07 & \textbf{65.78} & 60.14\\ 
        \end{tabular}
}

\newcommand{\bbb}{% Just for this example
  \begin{tabular}{p{0.9cm}<{\centering}|p{0.6cm}<{\centering}p{0.6cm}<{\centering}p{0.6cm}<{\centering}}

          & AUC & F1 & OA\\
        \midrule
        \midrule
        1 & 62.41 & 58.32 & 56.06\Bstrut\\ 
        2 & \textbf{67.48} & \textbf{68.53} & \textbf{63.29}\Bstrut\\ 
        4 & 64.32 & 63.51 & 61.56\\ 
        6 & 63.71 & 60.46 & 59.53\\
        \end{tabular}
}

\newcommand{\ccc}{% Just for this example
\begin{tabular}{p{0.9cm}<{\centering}|p{0.6cm}<{\centering}p{0.6cm}<{\centering}p{0.6cm}<{\centering}}

          & AUC & F1 & OA\\
        \midrule
        \midrule
        2 & 63.56 & 61.47 & 60.75\\
        4 & 67.48 & \textbf{68.53} & 63.29\Bstrut\\ 
        6 & \textbf{68.12} & 67.40 & \textbf{64.18}\Bstrut\\ 
        8 & 66.24 & 65.83 & 62.61\\ 
        \end{tabular}
}

\newcommand{\ddd}{% Just for this example
\begin{tabular}{p{0.9cm}<{\centering}|p{0.6cm}<{\centering}p{0.6cm}<{\centering}p{0.6cm}<{\centering}}

          & AUC & F1 & OA\\
        \midrule
        \midrule
        768 & 65.18 & 65.86 & 63.79\\
        1536 & 67.46 & 66.51 & 63.38\Bstrut\\ 
        3072 & 68.12 & 67.40 & \textbf{64.18}\Bstrut\\ 
        4096 & \textbf{68.65} & \textbf{67.68} & 63.87\\ 
        \end{tabular}
}

\begin{table*}%
\begin{adjustwidth}{-0.5cm}{-0.5cm}
\caption{\textbf{Ablations on the ANDT design.} We show AUC, F1 score, and overall accuracy (OA) of several Transformer designs with different configurations. The best accuracies are shown in \textbf{bold}.}
  \centering
  \subfloat[][\textbf{Patch size.} The model with $16 \times 16$ exhibits superior performance and effectively preserves the spatiotemporal information of the input video.]{\aaa}\hspace{2mm}
  \subfloat[][\textbf{Number of Transformer layers.} The network with 2 layers achieves a better performance and also has relatively small computational complexity.]{\bbb}\hspace{2mm}
  \subfloat[][\textbf{Number of attention heads.} The model with 6 attention heads has a outstanding performance and is able to learn long-term temporal features.]{\ccc}\hspace{2mm}
  \subfloat[][\textbf{MLP size.} The MLP with the size of 4096 achieves a better performance. Larger size MLP improves the information capacity of spatiotemporal features.]{\ddd}
  \label{tab:ablation}%
\end{adjustwidth}
\end{table*}

\begin{table}[]
\footnotesize
\centering
\begin{threeparttable}
\caption{\textbf{Prediction \emph{vs.} Reconstruction.} We show numerical results of three different anomaly detection strategies. The best results are shown in \textbf{bold}.}
\label{Prediction}
\centering
% \begin{tabular}{p{1.7cm}|*{3}{p{1.7cm}<{\centering}}}
\begin{tabular}{p{2.0cm}|p{0.5cm}<{\centering}p{0.6cm}<{\centering}p{0.8cm}<{\centering}p{1.0cm}<{\centering}p{0.5cm}<{\centering}p{0.5cm}<{\centering}}

\toprule
\textbf{Model} & AUC & Recall & Precision & F1 score & OA & $\Delta_s$\\
\midrule
\midrule

Reconstruction-1\tnote{1} & 62.1 & 64.9 & 60.3 & 62.5 & 60.6 & 0.16\\ 

Reconstruction-6\tnote{2} & 66.7 & 64.4 & 62.5 & 63.4 & 63.1 & 0.19\\ 

Prediction-1\tnote{3} & \textbf{68.7} & \textbf{68.4} & \textbf{66.9} & \textbf{67.7} & \textbf{65.9} & \textbf{0.25}\\

\bottomrule
\end{tabular}

\begin{tablenotes}
    \item[1] Reconstruction-1 is the strategy of inputting $1$ frame and reconstructing itself.
    \item[2] Reconstruction-6 is the strategy of inputting $6$ consecutive frames and reconstructing themselves.
    \item[3] Prediction-1 is the strategy of inputting $6$ consecutive frames and predicting the next frame.
\end{tablenotes}

\end{threeparttable}
\end{table}

\section{Experiments} \label{sec:Experiments}

In this section, we present our experimental results. In Section~\ref{sec:Dataset}, We introduce the datasets used in experiments. Evaluation metrics are introduced in Section~\ref{sec:Metrics}. Next, several ablation studies are conducted to investigate the effectiveness of our method, and we report their results in Section~\ref{Ablation}. Moreover, in Section~\ref{results}, we provide a benchmark on Drone-Anomaly dataset for anomaly detection in aerial videos by extensively validating existing methods, and we compare our method with these baseline models. In section~\ref{sec:results_AU-AIR-Anomaly}, we assess the performance of our method on AU-AIR-Anomaly dataset and compare our method with other competitors. Finally, we visualize the learned features of our method in Section~\ref{sec:visualization}.

\subsection{Dataset}
\label{sec:Dataset}
To evaluate the performance of our method, we use not only our Drone-Anomaly dataset but also the AU-AIR-Anomaly dataset~\cite{bozcan2021context}. A statistic of the two datasets can be found in Table~\ref{dataset_details}.

\subsection{Evaluation Metrics}
\label{sec:Metrics}

The receiver operation characteristic (ROC) is a popular evaluation matrix in anomaly detection, and it is calculated by gradually changing the threshold. In addition, we also use area under curve (AUC) for the performance evaluation. We leverage a strategy to determine a threshold that is used to calculate recall, precision, F1 score, and overall accuracy (OA). Specifically, we feed the training set into the trained model to obtain reconstruction errors for all training samples. The threshold is determined as the sum of the mean value and the standard deviation value of the reconstruction errors. We note that AUC is the primary metric, as it can comprehensively evaluate the performance of a method.

\subsection{Competitors}
\label{sec:competitors}

We compare our network with several state-of-the-art anomaly detection models.

\begin{itemize}
    \item CAE~\cite{masci2011stacked}: CAE (convolutional autoencoder) aims to leverage the convolutional encoder to map the input frames into a latent space to learn features. A convolutional decoder is then employed to reconstruct a frame based on the learned features. Its reconstruction error is used for detecting anomalies. 
    \item CVAE~\cite{nguyen2019anomaly}: CVAE (convolutional variational autoencoder) introduces a regularization into the representation space. It utilizes a prior distribution over the latent space to encode normal instances. This prevents the overfitting problem and enables the generation of meaningful frames for anomaly detection. 
    \item adVAE~\cite{wang2020advae}: adVAE (self-adversarial variational autoencoder) assumes that both anomalous and normal prior distributions are Gaussian. It utilizes a self-adversarial mechanism that adds discrimination training objectives to the encoder and decoder.
    \item GANomaly~\cite{akcay2018ganomaly}: GANomaly leverages a conditional generative adversarial network (GAN) to learn high-dimensional visual representations. It employs an encoder-decoder-encoder architecture in the generator network to enable the model to learn discriminative features of normality.
    \item Skip-GANomaly~\cite{akccay2019skip}: Skip-GANomaly employs a convolutional encoder-decoder architecture with skip connections to thoroughly capture the multi-scale distribution of normality.
\begin{table}[]
\footnotesize
\centering
\caption{\textbf{Number of input frames.} We report the performance of our model with a variant number of input frames. The best accuracies are shown in \textbf{bold}.}
\label{input frames}
\centering
% \begin{tabular}{p{1.7cm}|*{3}{p{1.7cm}<{\centering}}}
\begin{tabular}{p{0.5cm}<{\centering}|p{0.6cm}<{\centering}p{0.6cm}<{\centering}p{0.75cm}<{\centering}p{1.0cm}<{\centering}p{0.6cm}<{\centering}}

\toprule
  & AUC & Recall & Precision & F1 score & OA\\
\midrule
\midrule

2 & 63.7 & 68.0 & 58.9 & 63.1 & 62.5\\
4 & 67.4 & 64.5 & 69.7 & 67.0 & 65.1\Bstrut\\ 
6 & \textbf{68.7} & 68.4 & 66.9 & \textbf{67.7} & \textbf{65.9}\Bstrut\\ 
8 & 67.1 & 63.2 & \textbf{71.5} & 67.0 & 65.4\Bstrut\\ 
10 & 65.8 & 64.9 & 65.4 & 65.2 & 63.0\Bstrut\\ 
12 & 64.0 & \textbf{70.3} & 60.4 & 65.0 & 62.2\\

\bottomrule
\end{tabular}
\end{table}

    \item MemAE~\cite{gong2019memorizing}: MemAE (memory-augmented autoencoder) introduces a memory block between the encoder and decoder. It records prototypical normal patterns optimally and efficiently by the proposed sparse addressing strategy. 
    \item MNAD~\cite{park2020learning}: MNAD (memory-guided normality for anomaly detection) uses a memory module to record multiple prototypes that represent diverse representations of normalities for unsupervised anomaly detection. 

\begin{table*}[t]
\caption{\textbf{Comparing our approach against other methods.} We compare our ANDT with other competitors on highway, crossroads, bike roundabout, and vehicle roundabout scenes. The best accuracies are shown in \textbf{bold}.}
\label{results_part1}
\centering

% \begin{tabular}{@{}c|ccccc|ccccc|ccccc|ccccc|@{}}
\begin{adjustwidth}{-0.48cm}{0cm}
\begin{tabular}{r|*{5}{p{0.15cm}<{\centering}}|*{5}{p{0.15cm}<{\centering}}|*{5}{p{0.15cm}<{\centering}}|*{5}{p{0.15cm}<{\centering}}}
\toprule
\multirow{2}{*}{Model} & \multicolumn{5}{c|}{\textbf{Highway}}                                       & \multicolumn{5}{c|}{\textbf{Crossroads}} & \multicolumn{5}{c|}{\textbf{Bike roundabout}} & \multicolumn{5}{c}{\textbf{Vehicle roundabout}}                              \\ [4pt] \cline{2-21} \Tstrut
                       & \multicolumn{1}{l}{\rotatebox{75}{AUC}} & \multicolumn{1}{l}{\rotatebox{75}{Recall}} & \multicolumn{1}{l}{\rotatebox{75}{Precision}} & \multicolumn{1}{l}{\rotatebox{75}{F1 score}} & \multicolumn{1}{l|}{\rotatebox{0}{OA}}& \multicolumn{1}{l}{\rotatebox{75}{AUC}} & \multicolumn{1}{l}{\rotatebox{75}{Recall}} & \multicolumn{1}{l}{\rotatebox{75}{Precision   }} & \multicolumn{1}{l}{\rotatebox{75}{F1 score}} & \multicolumn{1}{l|}{\rotatebox{0}{OA }}& \multicolumn{1}{l}{\rotatebox{75}{AUC}} & \multicolumn{1}{l}{\rotatebox{75}{Recall}} & \multicolumn{1}{l}{\rotatebox{75}{Precision}} & \multicolumn{1}{l}{\rotatebox{75}{F1 score}} & \multicolumn{1}{l|}{\rotatebox{0}{OA }}& \multicolumn{1}{l}{\rotatebox{75}{AUC}} & \multicolumn{1}{l}{\rotatebox{75}{Recall}} & \multicolumn{1}{l}{\rotatebox{75}{Precision}} & \multicolumn{1}{l}{\rotatebox{75}{F1 score}} & \multicolumn{1}{l}{\rotatebox{0}{OA }} \\ 
\midrule
\midrule

CAE~\cite{masci2011stacked} & 58.3 & 60.4 & 58.8 & 59.6 & 57.1 & 57.7 & 60.7 & 61.3 & 61.0 & 60.3 & 59.4 & 57.7 & 59.0 & 58.3 & 58.8 & 60.9 & 58.9 & 56.5 & 57.7 & 58.4  \\  [2pt]

CVAE~\cite{nguyen2019anomaly} & 61.7 & 64.1 & 63.4 & 63.7 & 61.0 & 62.4 & 61.5 & 61.8 & 61.7 & 59.7 & 76.5 & 68.8 & 73.4 & 71.0 & 68.7 & 57.6 & 58.4 & 57.6 & 58.0 & 56.4   \\  [2pt]

adVAE~\cite{wang2020advae} & 61.1 & 59.7 & 60.3 & 60.0 & 59.1 & 56.1 & 56.9 & 54.8 & 55.8 & 56.5 & 72.8 & 71.8 & 75.9 & 73.8 & 69.4 & 55.1 & 54.4 & 52.9 & 53.6 & 54.1 \\  [2pt]

GANomaly~\cite{akcay2018ganomaly} & 62.7 & 65.1 & 62.9 & 64.0 & 61.5 & 58.9 & 58.5 & 57.2 & 57.9 & 59.2 & 71.7 & 70.2 & 77.5 & 73.7 & 69.3 & 55.1 & 58.6 & 55.7 & 57.1 & 54.0  \\  [2pt]

Skip-GAN~\cite{akccay2019skip} & 64.8 & 63.7 & 66.7 & 65.2 & 64.6 & 59.3 & 60.3 & 60.6 & 60.4 & 62.1 & 77.7 & 73.5 & 74.3 & 73.9 & 67.7 & 58.5 & 59.3 & 62.8 & 61.0 & 57.1 \\  [2pt]

MemAE~\cite{gong2019memorizing} & 67.2 & 67.3 & \textbf{68.2} & 67.7 & 66.1 & 64.1 & \textbf{63.8} & 63.3 & 63.6 & 64.5 & 79.5 & 74.8 & 73.4 & 74.1 & 75.2 & \textbf{64.1} & \textbf{60.8} & 59.0 & 59.9 & 59.2 \\  [2pt]

MNAD~\cite{park2020learning} & 66.9 & 65.9 & 66.5 & 66.2 & 65.7 & 56.6 & 57.2 & 59.4 & 58.3 & 55.2 & 77.4 & 72.4 & 75.2 & 73.8 & 69.8 & 61.9 & 57.9 & 61.6 & 59.7 & 59.4\\  [2pt]

%HF$^2$-VAD~\cite{ristea2022self} & 67.4 & 62.6 & 68.1 & 65.2 & \textbf{66.3} & 63.4 & 62.1 & 60.7 & 61.4 & 60.8 & 81.3 & \textbf{79.8} & 77.6 & 78.7 & 77.0 & 63.8 & \textbf{61.5} & \textbf{66.7} & \textbf{64.0} & \textbf{61.4}\\  [2pt]

MKD~\cite{salehi2021multiresolution} & 64.3 & 62.8 & 65.3 & 64.0 & 63.9 & 63.5 & 63.4 & 61.2 & 62.3 & 63.7 & 74.8 & 70.6 & 75.1 & 72.8 & 73.2 & 62.7 & 59.7 & 63.7 & 61.6 & 58.7\\  [2pt]

SSPCAB~\cite{ristea2022self} & 67.8 & 67.5 & 69.7 & \textbf{68.6} & \textbf{66.3} & 60.4 & 60.7 & 61.9 & 61.3 & 60.4 & 76.8 & 74.6 & 76.0 & 75.3 & 70.4 & 62.3 & 59.7 & 63.8 & \textbf{61.7} & \textbf{60.4}\\  [2pt]

\hline 

ANDT  & \textbf{68.7} & \textbf{68.4} & 66.9 & 67.7 & 65.9 & \textbf{65.2} & 63.1 & \textbf{66.3} & \textbf{64.6} & \textbf{65.8} & \textbf{82.2} & \textbf{78.5} & \textbf{79.0} & \textbf{78.8} & \textbf{76.7} & 61.3 & 57.8 & \textbf{64.1} & 60.8 & 58.0\Tstrut\\

\bottomrule
\end{tabular}
\end{adjustwidth}
\end{table*}

\begin{table*}[t]
\caption{\textbf{Comparing our approach against other methods.} We compare our ANDT with other competitors on railway inspection, solar panel inspection, and farmland inspection scenes. The best accuracies are shown in \textbf{bold}.}
\label{results_part2}
\centering

% \begin{tabular}{@{}r|ccccc|ccccc|ccccc@{}}
\begin{tabular}{r|*{5}{p{0.15cm}<{\centering}}|*{5}{p{0.15cm}<{\centering}}|*{5}{p{0.15cm}<{\centering}}}
\toprule
\multirow{2}{*}{Model} & \multicolumn{5}{c|}{\textbf{Railway inspection}}                                       & \multicolumn{5}{c|}{\textbf{Solar panel inspection}} & \multicolumn{5}{c}{\textbf{Farmland inspection}}                              \\ [4pt] \cline{2-16} \Tstrut
                       & \multicolumn{1}{l}{\rotatebox{75}{AUC}} & \multicolumn{1}{l}{\rotatebox{75}{Recall}} & \multicolumn{1}{l}{\rotatebox{75}{Precision}} & \multicolumn{1}{l}{\rotatebox{75}{F1 score}} & \multicolumn{1}{l|}{\rotatebox{0}{OA}}& \multicolumn{1}{l}{\rotatebox{75}{AUC}} & \multicolumn{1}{l}{\rotatebox{75}{Recall}} & \multicolumn{1}{l}{\rotatebox{75}{Precision  }} & \multicolumn{1}{l}{\rotatebox{75}{F1 score}} & \multicolumn{1}{l|}{\rotatebox{0}{OA }}& \multicolumn{1}{l}{\rotatebox{75}{AUC}} & \multicolumn{1}{l}{\rotatebox{75}{Recall}} & \multicolumn{1}{l}{\rotatebox{75}{Precision}} & \multicolumn{1}{l}{\rotatebox{75}{F1 score}} & \multicolumn{1}{l}{\rotatebox{0}{OA }} \\ 
\midrule
\midrule

CAE~\cite{masci2011stacked} & 61.2 & 59.7 & 54.8 & 57.1 & 56.7 & 62.9 & \textbf{62.7} & 65.3 & \textbf{64.0} & 60.2 & 77.1 & 79.2 & 72.6 & 75.8 & 74.5   \\  [2pt]

CVAE~\cite{nguyen2019anomaly} & 59.1 & \textbf{62.8} & \textbf{64.7} & 62.2 & 59.3 & 57.5 & 57.3 & 57.1 & 57.2 & 58.4 & 78.4 & \textbf{80.7} & 77.1 & \textbf{78.9} & \textbf{75.7} \\  [2pt]

adVAE~\cite{wang2020advae} & 62.1 & 56.2 & 57.9 & 57.1 & 56.4 & \textbf{66.1} & 58.6 & 60.9 & 59.8 & 60.5 & 73.8 & 77.9 & 76.6 & 77.3 & 72.6 \\  [2pt]

\quad GANomaly~\cite{akcay2018ganomaly} & 61.7 & 55.7 & 56.2 & 56.0 & 53.8 & 64.6 & 59.1 & 63.7 & 61.3 & 57.3 & 77.1 & 74.0 & 73.2 & 73.6 & 75.5\\  [2pt]

Skip-GAN~\cite{akccay2019skip} & \textbf{65.8} & 60.7 & 64.6 & \textbf{62.6} & 60.3 & 65.7 & 58.8 & 57.5 & 58.1 & 60.2 & 71.7 & 75.4 & 73.3 & 74.3 & 72.6\\  [2pt]

MemAE~\cite{gong2019memorizing} & 58.9 & 58.0 & 58.4 & 58.2 & 58.0 & 65.8 & 62.1 & 57.6 & 59.8 & 57.7 & 74.1 & 79.7 & \textbf{77.7} & 78.7 & 74.4 \\  [2pt]

MNAD~\cite{park2020learning}  & 58.0 & 61.3 & 56.1 & 58.6 & 57.1 & 64.7 & 58.6 & 58.0 & 58.3 & 59.6 & 78.6 & 78.5 & 74.2 & 76.3 & 74.5\\  [2pt]

% ANDT  & \textcolor{blue}{59.4} & \textcolor{blue}{60.7} & \textcolor{blue}{61.3} & \textcolor{blue}{61.0} & \textcolor{blue}{57.4} & 64.2 & 61.2 & \textbf{66.0} & 63.5 & \textbf{60.8} & \textcolor{blue}{\textbf{79.5}} & \textcolor{blue}{76.9} & \textcolor{blue}{\textbf{77.6}} & \textcolor{blue}{77.2} & \textcolor{blue}{73.5}\Tstrut\\ 

MKD~\cite{salehi2021multiresolution} & 62.4 & 59.7 & 60.3 & 60.0 & \textbf{60.8} & 63.5 & 57.6 & 54.7 & 56.1 & 56.5 & 75.2 & 76.8 & 72.4 & 74.5 & 72.8\\  [2pt]

SSPCAB~\cite{ristea2022self} & 59.1 & 62.0 & 58.7 & 60.3 & 59.4 & 65.0 & 59.2 & 60.9 & 60.0 & 58.7 & 79.0 & 78.4 & 75.8 & 77.9 & 75.1\\  [2pt]

\hline 
ANDT  & 59.4 & 60.7 & 61.3 & 61.0 & 57.4 & 64.2 & 61.2 & \textbf{66.0} & 63.5 & \textbf{60.8} & \textbf{79.5} & 76.9 & 77.6 & 77.2 & 73.5\Tstrut\\ 

\bottomrule
\end{tabular}
\end{table*}

    \item
    MKD~\cite{salehi2021multiresolution}: MKD (multiresolution knowledge distillation for anomaly detection) proposes to distill the knowledge of a pre-trained expert network into another more compact network to concentrate solely on discriminative features that are helpful in distinguishing normality and anomaly.

    \item
    SSPCAB~\cite{ristea2022self}: SSPCAB (self-supervised predictive convolutional attentive block) uses a convolutional layer with dilated filters, where the center area of the receptive field is masked. The block learns to reconstruct the masked area using contextual information. It can be incorporated into various existing models. In this paper, we equip it on the MNAD~\cite{park2020learning} model, which is still denoted SSPCAB.
\end{itemize}

\subsection{Ablation Studies}
\label{Ablation}

We present a series of ablations for evaluating the effectiveness of our model. All of them are conducted on the \emph{highway} scene with the most number of training and test frames.

% \begin{table}[]
% \footnotesize
% \centering
% \caption{We report several evaluation results of our model and other competitors.The best accuracies are shown in \textbf{bold}.}
% \label{tab:results}
% \centering
% % \begin{tabular}{p{1.7cm}|*{3}{p{1.7cm}<{\centering}}}
% \begin{tabular}{p{2.5cm}<{\centering}|p{0.6cm}<{\centering}p{0.6cm}<{\centering}p{0.75cm}<{\centering}p{1.0cm}<{\centering}p{0.6cm}<{\centering}}

% \toprule
%   & AUC & Recall & Precision & F1 score & OA\\
% \midrule
% \midrule

% No Pos. Emb. & 75.36 & 72.77 & \textbf{82.30} & 77.24 & 71.98\\
% 1-D Pos. Emb. & 77.84 & 94.84 & 70.43 & 80.83 & 71.85\Bstrut\\ 
% 2-D Pos. Emb. & 77.74 & 89.55 & 70.28 & 78.76 & 68.43\Bstrut\\ 
% 3-D Pos. Emb. & 73.38 & 75.75 & 74.14 & 74.94 & 66.89\\ 

% \bottomrule
% \end{tabular}
% \end{table}

\textbf{Model design.} In the course of experiments, we find that the design of the Transformer encoder matters. Hence, we want to investigate different configurations and figure out optimal settings. Concretely, the following hyperparameters are taken into account: patch size, number of Transformer layers, number of attention heads, and MLP size. From Table~\ref{tab:ablation}\textcolor{red}{a}, it can be observed that the model with a patch size of $16 \times 16$ achieves better comprehensive performance. The patch size is actually associated with the extent to which the model excavates inner information in patches and spatiotemporal relations among patches. In Table~\ref{tab:ablation}\textcolor{red}{b} and~\ref{tab:ablation}\textcolor{red}{c}, we focus on self-attention and find that using 2 Transformer layers and 6 attention heads exhibits superior performance. Multi-head self-attention enables the model to integrate multiple temporal information from different representation patches. And the small number of Transformer layers ensures a relatively small computational complexity. Finally, MLP size determines the size of the output spatiotemporal feature of the Transformer encoder. In Table~\ref{tab:ablation}\textcolor{red}{d}, we can see that an MLP with a size of 4096 brings good results to our model, which could be caused by the improved information capacity of the spatiotemporal feature.

\begin{figure*}[]
	\centering
 	\includegraphics[width = 0.97\linewidth]{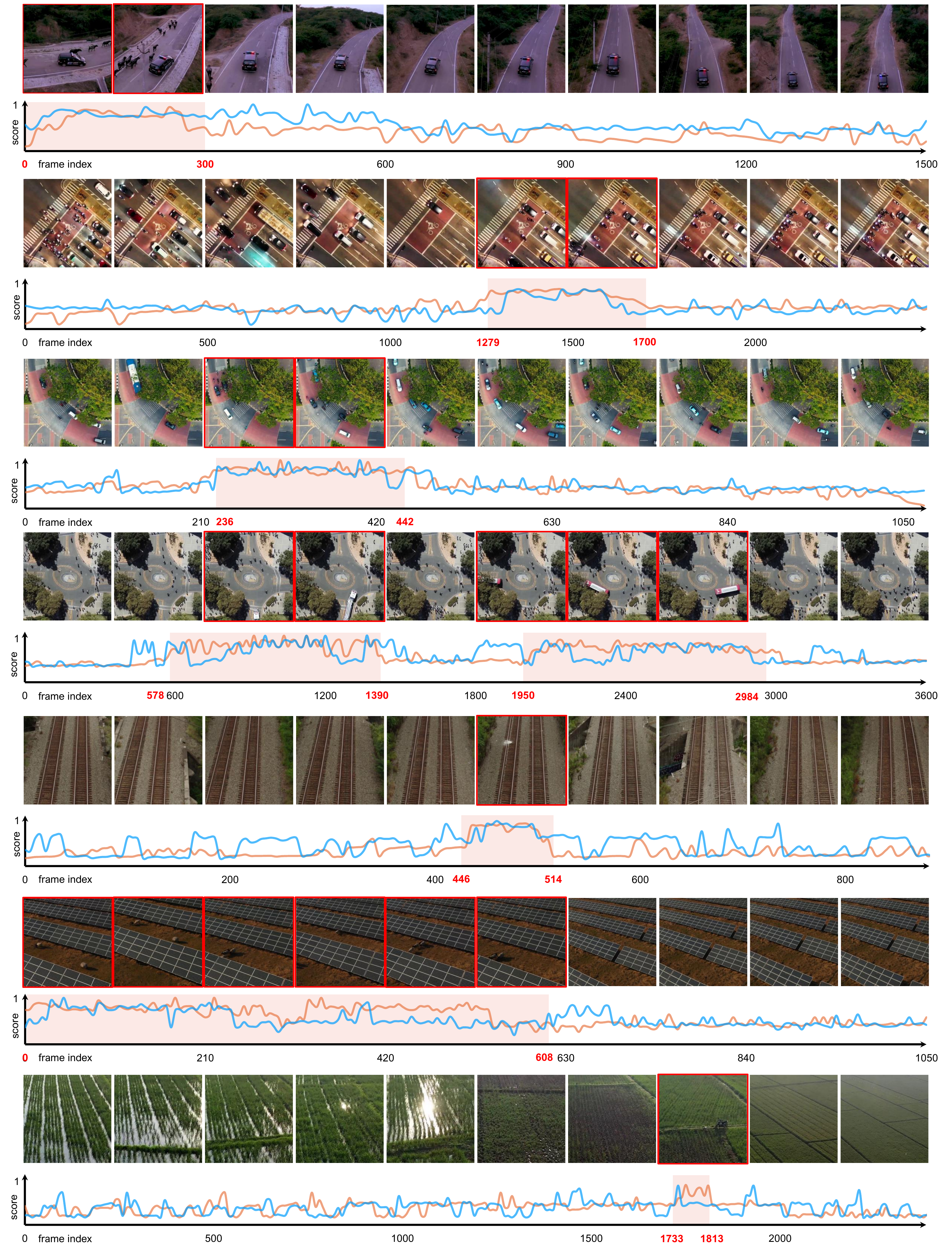}
	\caption{\textbf{Visualization of anomaly detection results of our method and a main competitor.} We show frame-level anomaly scores (\textcolor[RGB]{233,123,73}{orange} curves indicate ANDT and \textcolor[RGB]{3,158,253}{blue} curves denote MemAE). Ten frames of each video are shown, and anomalous frames are marked with \textcolor{red}{red} borders. \textcolor{red}{Red} rectangles are ground truth data. A demo video is available at \url{https://youtu.be/ancczYryOBY}.}
	\label{vis_results}
\end{figure*}

\textbf{Prediction \emph{vs.} Reconstruction.} In our network, future frame prediction is an important strategy to learn temporal dependencies for effectively detecting anomalies. To evaluate how it affects the performance, we compare our prediction-based framework with a commonly used reconstruction-based methodology~\cite{an2015variational,masci2011stacked,fowler2011anomaly,zong2018deep,nguyen2019anomaly,wang2020advae,akcay2018ganomaly,akccay2019skip,gong2019memorizing,park2020learning}. More specifically, with the same network architecture, we consider the following models: (1) inputting 1 frame, reconstructing itself; (2) inputting 6 consecutive frames, reconstructing themselves; (3) inputting 6 consecutive frames, predicting the next frame (i.e., the proposed method). We first report results of these models in the five evaluation metrics. Then, we calculate the difference between the average anomaly score of normal frames and that of abnormal frames, represented by $\Delta_s$. The network with a relatively large $\Delta_s$ is more capable of distinguishing abnormal frames from normal frames. All results are shown in Table~\ref{Prediction}. It can be seen that the prediction-based framework can achieve better results in AUC, Recall, F1 score, OA, and $\Delta_s$. 
%Therefore, the difference of temporal features between the normality and anomaly in aerial videos is significant and can be effectively utilized for detecting anomalies.  

\begin{table}[t]
\caption{\textbf{AUC results of different kinds of anomalies in Crossroads.} We offer AUC results of two kinds of anomalies in crossroads. The best accuracies are shown in \textbf{bold}.}
\label{results_crossroads}
\centering
\begin{tabular}{r|*{2}{p{2cm}<{\centering}}}
\toprule
\multirow{2}{*}{Model} & \multicolumn{2}{c}{\textbf{Crossroads}} \\ [4pt] \cline{2-3} \Tstrut 

&

\multicolumn{1}{l}{\emph{person-related}} & \multicolumn{1}{l}{\emph{vehicle-related}} \\

\midrule
\midrule

CAE~\cite{masci2011stacked} & 61.8 & 55.0  \\  [2pt]
CVAE~\cite{nguyen2019anomaly} & 59.9 & 64.1 \\  [2pt]
adVAE~\cite{wang2020advae} & 57.2 & 55.4 \\  [2pt]
\quad GANomaly~\cite{akcay2018ganomaly} & 52.0 & 63.5 \\ [2pt]
Skip-GAN~\cite{akccay2019skip} & 56.4 & 61.2\\  [2pt]
MemAE~\cite{gong2019memorizing} & 64.7 & 63.7 \\  [2pt]
MNAD~\cite{park2020learning}  & 57.3 & 56.1 \\  [2pt]
MKD~\cite{salehi2021multiresolution}  & 62.7 & 64.3 \\  [2pt]
SSPCAB~\cite{ristea2022self}  & 58.7 & 62.1 \\  [2pt]

\hline 
ANDT  & \textbf{65.8} & \textbf{64.8}\Tstrut\\ 

\bottomrule
\end{tabular}
\end{table}

\textbf{Number of input frames.} We further investigate how the number of input frames affects the performance of our method. We evaluate the performance of ANDT with a variant number of input frames. The results are reported in Table~\ref{input frames}. We can see that the method with $6$ input frames exhibits superior comprehensive performance. The performance of our model gradually gets better as the number of input frames goes from $2$ to $6$ and then degrades with more input frames. This observation demonstrates that a few frames are not enough for modeling temporal context, but too many input frames bring a deteriorated performance.

\subsection{Results on the Drone-Anomaly Dataset} \label{results}

We evaluate various baseline models on all scenes in our Drone-Anomaly dataset with standard evaluation protocols and offer a benchmark. The results are reported in Table~\ref{results_part1} and~\ref{results_part2}. Also, we compare the proposed model with other competitors. 

\emph{Highway}. This scene presents various kinds of anomalous events, e.g., a cow herd walking on the street, an accidental car collision, and a road section covered by sand and dust. These different anomalous events make this scenario very challenging. Comparing to other competitors, our method achieves the best results in AUC (68.7\%), and Recall (68.4\%). The main competitor in this scene is MemAE that also exhibits very good results in some metrics. However, its accuracy in AUC is relatively a bit low. Our method demonstrates the capability of detecting different anomalous events and even presents better performance than memory-based methods, such as MemAE and MNAD, that are specially designed to deal with various anomalies.

\emph{Crossroads}. This scene focuses on distinguishing various anomalous behaviors of vehicles and persons, such as persons crossing the road irregularly and vehicles moving backward. In this scene, capturing temporal dynamics of persons and vehicles on the road is critical for identifying their anomalous behaviors. From the reported results in Table~\ref{results_part2}, our method achieves the best results in AUC (65.2\%), Precision (66.3\%), F1 score (64.6\%), and OA (65.8\%). This is mainly because the Transformer encoder of our approach is able to effectively model long-term temporal relations for distinguishing anomalous moving directions of persons or vehicles. We visualize the prediction of our method on a video clip of this scenario in Fig.~\ref{vis_results} (see the third row), in which an anomalous event is that a person crosses the road not following the rule. We can observe that the traffic is hindered by the person crossing the road irregularly. In this case, dynamically sensing traffic speed is crucial for the successful detection of anomalous events. The numerical results demonstrate the effectiveness of our model. For evaluating the performance of detecting different kinds of anomalous events, we group anomalies into two categories: person-related anomaly and vehicle-related anomaly. The AUC results of each anomalous event are reported in Table~\ref{results_crossroads}. Compared to other methods, our approach achieves the best AUC results in both two kinds of anomalies.

\begin{table}[t]
\footnotesize
\centering
\caption{\textbf{{Comparing our approach against other methods on the AU-AIR-Anomaly dataset.}} We compare our ANDT with other competitors on AU-AIR dataset. The best accuracies are shown in \textbf{bold}.}
\label{AU-AIR}
\centering
% \begin{tabular}{p{1.7cm}|*{3}{p{1.7cm}<{\centering}}}
\begin{tabular}{r|p{0.6cm}<{\centering}p{0.6cm}<{\centering}p{0.75cm}<{\centering}p{1.0cm}<{\centering}p{0.6cm}<{\centering}}

\toprule
Model  & AUC & Recall & Precision & F1 score & OA\\
\midrule
\midrule

CAE~\cite{masci2011stacked} & 69.3 & 70.2 & 64.7 & 67.3 & 66.4\\  [2pt]
CVAE~\cite{nguyen2019anomaly} & 70.8 & 63.7 & 72.1 & 67.6 & 67.1\Bstrut\\  [2pt]
adVAE~\cite{wang2020advae} & 72.2 & 70.7 & 74.9 & 72.7 & 70.6\Bstrut\\  [2pt]
GANomaly~\cite{akcay2018ganomaly} & 70.4 & 73.6 & 61.8 & 67.2 & 72.8\Bstrut\\  [2pt]
Skip-GAN~\cite{akccay2019skip} & 74.8 & 60.8 & 84.1 & 70.6 & 72.1\Bstrut\\  [2pt]
MemAE~\cite{gong2019memorizing} & 81.4 & \textbf{87.6} & 74.8 & 80.7 & \textbf{82.4} \\ [2pt]
MNAD~\cite{park2020learning} & 78.4 & 76.9 & 79.4 & 78.1 & 76.2\\ [2pt]
MKD~\cite{salehi2021multiresolution} & 76.8 & 83.7 & 79.6 & 81.6 & 79.5\\ [2pt]
SSPCAB~\cite{ristea2022self} & 79.6 & 77.4 & 80.4 & 78.9 & 78.3\\ [2pt]

\hline 
ANDT & \textbf{86.7} & 80.7 & \textbf{84.9} & \textbf{82.7} & 82.0\Tstrut\\

\bottomrule
\end{tabular}
\end{table}

\emph{Bike roundabout}. Only one type of anomaly, i.e., moving vehicle on the bike roundabout, is presented in this scene. However, more than one abnormal event may be present in the test video sequence. This scenario can verify whether a method is able to continuously detect all anomalous events in a test sequence. Our method exhibits superior performance. We also observe that memory-based methods have poor performance. The reason for this may be that some feature representations of abnormal video frames misidentified as normality are memorized in the memory space, which deteriorates the performance of these models in recognizing subsequent anomalous frames.

\emph{Vehicle roundabout}. Various anomalous events, such as traffic congestion and people crossing the road irregularly, are present in this scene. Memory-based and GAN-based methods, namely Skip-GAN, MemAE, and MNAD, show superior performance in this scene. Our model suffers from insufficient training data and performs relatively poor.

\emph{Railway inspection}. This scene presents only one kind of anomaly, i.e., obstacles on the railway. Determining the existence of obstacles on the railway is vital in practical applications. From the results in Table~\ref{results_part2}, there is no dominant method. The reason might be the insufficient training data (only $400$ frames are available for training) cannot ensure that these models learn strong feature representations of normality. 

%Moreover, the anomaly has not demonstrated a dynamical change in the video sequence. Our approach is also not outstanding for detecting this anomaly.

\emph{Solar panel inspection}. Two anomalies, unknown objects/animals and panel defects, appear in this scene. Our model achieves the best accuracies in Precision (66.0\%) and OA (60.8\%), and provides relatively satisfactory results in this scenario. 

\emph{Farmland inspection}. One type of anomaly, i.e., unidentified vehicles, exists in this scene. Searching anomalous objects is the goal in this scene. From experimental results, our network achieves the best accuracies in AUC (79.5\%), and exhibits superior performance in searching anomalous objects.

In summary, our model exhibits superior performance in multiple scenes, including \emph{highway}, \emph{crossroads}, \emph{bike roundabout}, and \emph{farmland inspection}, in which many anomalous events with temporal dynamics exist. Specifically, in the \emph{highway} scene, our method presents a better performance of detecting different anomalies than memory-based methods, i.e., MemAE and MNAD, which are specially designed to deal with various anomalies. This is because the global temporal receptive field enables our model to learn discriminative temporal representations of normality, which is used to effectively detecting different anomalies.

\subsection{Results on the AU-AIR-Anomaly Dataset}
\label{sec:results_AU-AIR-Anomaly}

Further, we use AU-AIR-Anomaly dataset~\cite{bozcan2021context} to validate the performance of our approach and other methods. Due to the non-availability of public ground-truth labels for anomalies in the AU-AIR-Anomaly dataset, following~\cite{bozcan2021context}, we label four anomalous events: a car on a bike road, a person on a road, a parked van in front of a building and a bicycle on a road. We report numerical results in Table VI. As we can see, our model has a superb performance and achieves the best accuracies in AUC (86.7\%), Precision (84.9\%), and F1 score (82.7\%). The scene of this dataset is highly similar to crossroads in our Drone-Anomaly dataset. Our network still exhibits stable and superior performance, which demonstrates its good generalization ability across different datasets.

\subsection{Visualization of the Learned Features} 
\label{sec:visualization}

\begin{figure}[t]
	\centering
 	\includegraphics[width = 1\linewidth]{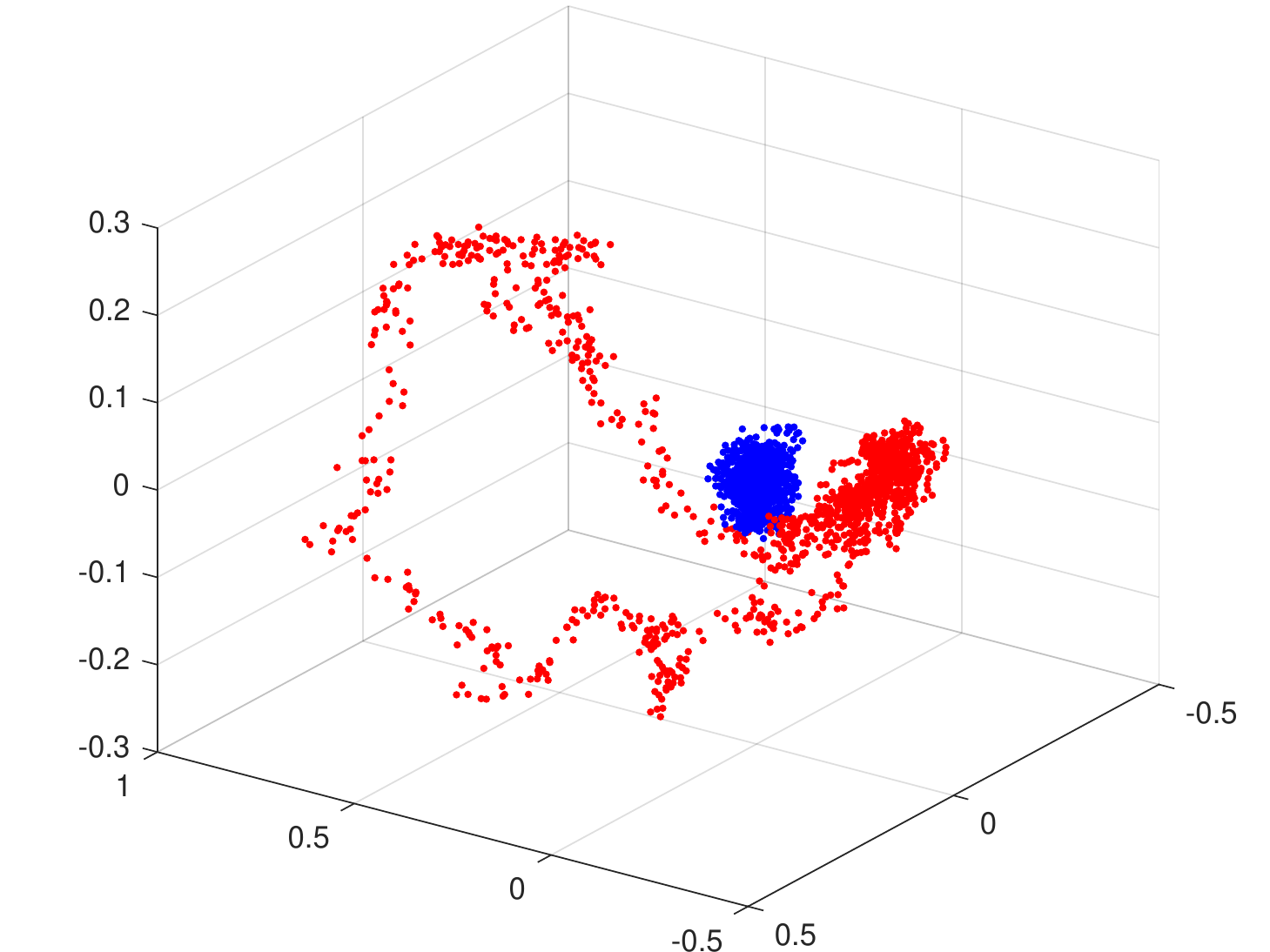}
	\caption{\textbf{Visualization of feature distribution.} We visualize the distribution of the learned spatiotemporal features from the Transformer encoder on the \emph{highway} scene. The features of normal frames are represented by \textcolor{blue}{blue} points, and features of anomalous frames are \textcolor{red}{red} points.}
	\label{feature_vis}
\end{figure}

\begin{figure}[t]
	\centering
 	\includegraphics[width = 1\linewidth]{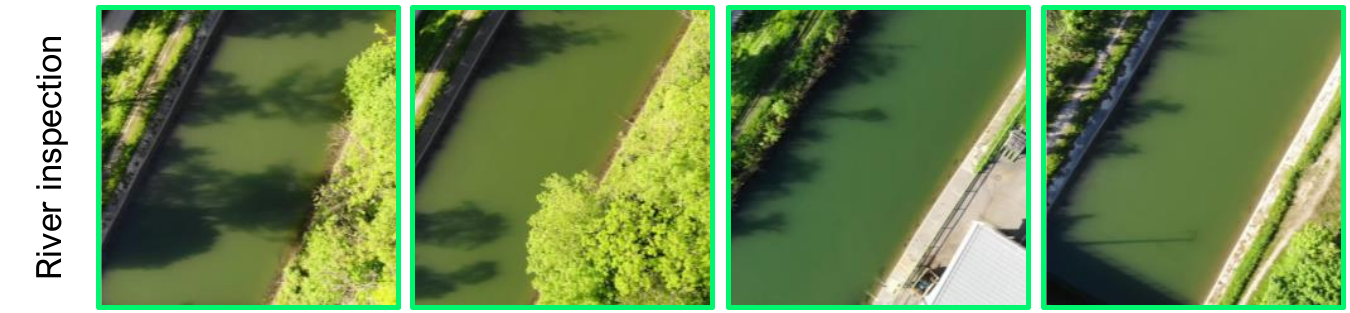}
	\caption{\textbf{Sample frames in the scene of river inspection.} We show four frames in this scene. All normal frames are marked with \textcolor{green}{green} borders.}
	\label{discuss_sample}
\end{figure}

We visualize in Fig.~\ref{feature_vis} the distribution Transformer features of some randomly chosen test samples on the \emph{crossroads} scene in the Drone-Anomaly dataset. We leverage principal component analysis (PCA) to reduce the dimension of the features to 3. From the figure, it can be seen that normal instances (blue points) are all concentrated in a relatively small area, while abnormal samples are far away from the blue cluster. This demonstrates that the spatiotemporal features learned by our model are very discriminative. 

% \begin{table}[]
% \footnotesize
% \centering
% \caption{\textbf{\textcolor{blue}{Mean Squared Reconstruction Rrror (MSRE) on different train and test sets.}} We compare MSRE results on different train and test sets in bike roundabout.}
% \label{results_disscussion}
% \centering
% % \begin{tabular}{p{1.7cm}|*{3}{p{1.7cm}<{\centering}}}
% \begin{tabular}{r|p{1.0cm}<{\centering}p{2.0cm}<{\centering}p{2.0cm}<{\centering}p{2.0cm}}

% \toprule
%   & Train data & Test data with anomalies & Test data without anomalies\\
% \midrule
% \midrule

% MSRE & 0.087 & 0.17 & 0.089\Tstrut\\

% \bottomrule
% \end{tabular}
% \end{table}

\subsection{Discussion} 

To verify whether our method raises too many false alarms in practical applications that do not contain any anomalies, we collect a new scene, i.e., river inspection, which does not contain anomalous events. We use a DJI drone to inspect a normal river and collect an aerial video for this validation. We show four sample frames of the test data in Fig.~\ref{discuss_sample}. We report mean squared reconstruction error (MSRE) values on training data and test data, and they are $MSRE_{tra}=0.076$ and $MSRE_{test}=0.078$. We can see that these two values are very close. Besides, we calculate false positive rate, $FPR=0.0041$, which is very low. These mean that in scenes without any anomalies, our model also works well.

\section{Conclusion} \label{Conclusion}

In this paper, we focus on detecting anomalous events in aerial videos. To this end, we create a new dataset, termed Drone-Anomaly, providing $37$ training video sequences and $22$ testing video sequences, covering $7$ real-world scenes, providing various anomalous events. Based on this dataset, we offer a benchmark for this task. Moreover, we present a new baseline model, ANDT, which treats a video as a sequence of tubelets and leverages a Transformer encoder to learn a spatiotemporal feature. Afterwards, a decoder is combined with the encoder for predicting the next frame based on the learned spatiotemporal representation. And we conduct extensive ablation studies for validating the effectiveness of our network. Moreover, we compare our model with other baselines. The experimental results demonstrate its outstanding performance. In the future, we will focus on spatiotemporally detecting anomalous events in aerial videos.

\ifCLASSOPTIONcaptionsoff
  \newpage
\fi

\bibliographystyle{IEEEtran}
\bibliography{references}

\end{document}